\documentclass{article}

\usepackage[nonatbib,final]{nips_2016}
\usepackage[numbers]{natbib}

\usepackage[utf8]{inputenc} 
\usepackage{hyperref}       
\usepackage{url}            
\usepackage{booktabs}       
\usepackage{amsfonts}       
\usepackage{nicefrac}       
\usepackage{microtype}      

\usepackage{color}
\usepackage{amsmath}
\usepackage{graphics}
\usepackage{epsfig}

\usepackage{caption}
\usepackage{subcaption}

\def\F{\mc{F}}

\def\Img{p}
\def\ImgSpace{\mc{P}}
\def\ImgOpt{p_*}
\def\Faces{\mc{P}^*}
\def\GImg{\Img_G}
\def\SetImgs{\mathfrak{p}}

\def\Emb{e}
\def\EmbSpace{\mc{E}}
\def\Pers{s}

\def\Reg{R}
\def\GReg{\Reg_G}
\def\RegTV{\Reg_{\rm TV}}
\def\RegM{\Reg_{\rm Mirror}}
\def\RegLP{\Reg_{\rm LP}}
\def\RegGauss{\Reg_{\rm Gauss}}
\def\wG{w_G}
\def\wtv{w_{\rm TV}}

\def\Act{a}
\def\NetF{\mc{F}}
\def\Metric{d}
\def\Prob{P}

\def\Av{v}
\def\Dev{\sigma}
\newcommand{\op}[1]{\hat{#1}}
\def\ShiftOp{\op{S}}
\def\MirrorOp{\op{F}_x}

\def\LPComponent{L}
\def\LPNorm{N_L}

\newcommand*\Laplace{\mathop{}\!\mathbin\bigtriangleup}

\def\R{\mathbb{R}}

\newcommand{\avr}[1]{\langle #1 \rangle}

\newcommand{\mc}[1]{\mathcal{#1}}

\newcommand{\Eq}[1]{Eq.~\eqref{#1}}

\newcommand{\Ref}[1]{Ref.~\citenum{#1}}

\newcommand{\Fig}[1]{Fig.~\ref{#1}}
\newcommand{\Figs}[1]{Figs.~\ref{#1}}

\newcommand{\Sec}[1]{Sec.~\ref{#1}}
\newcommand{\Secs}[1]{Secs.~\ref{#1}}
\newcommand{\App}[1]{Appendix~\ref{#1}}

\newcommand{\hairsp}{\hspace{1pt}}
\def \ie{{\mbox{\textit{i.\hairsp{}e.},~}}}

\newcommand{\pd}[2]{\frac{\partial #1}{\partial #2}}

\newcommand{\pld}[2]{{{\partial} #1}/{{\partial} #2}}

\newcommand{\argmin}{\mathop{\mathrm{arg\,min}}}
\usepackage[font=small,labelfont=bf]{caption}
\newcommand{\omt}[1]{}
\newcommand{\omtnips}[1]{#1}
\newcommand{\marginnote}[1]{}

\title{Inverting face embeddings with convolutional neural networks}

\author{
  Andrey~Zhmoginov \\
  Google Inc. \\
  1600 Amphitheatre Parkway \\
  Mountain View, CA 94043 \\
  \texttt{azhmogin@google.com} \\
  \And
  Mark~Sandler \\
  Google Inc. \\
  1600 Amphitheatre Parkway \\
  Mountain View, CA 94043 \\
  \texttt{sandler@google.com} \\
}

\begin{document}

\maketitle

\begin{abstract}
  Deep neural networks have dramatically advanced the state of the art for many areas of machine learning. Recently they have been shown
  to have a remarkable ability to generate highly complex visual artifacts such as images and text rather than simply recognize them.

  In this work we use neural networks to effectively invert low-dimensional face embeddings while producing
  realistically looking consistent images.
  Our contribution is twofold, first we show that a gradient ascent style approaches can be used
  to reproduce consistent images, with a help of a guiding image.
  Second, we demonstrate that we can train a separate neural network to effectively
  solve the minimization problem in one pass, and generate images in real-time.
  We then evaluate the loss imposed by using a neural network
  instead of the gradient descent by comparing the final values of the minimized loss function.
\end{abstract}


\section{Introduction}
\label{sec:introduction}

Deep neural networks are an extremely powerful tool for object recognition \cite{krizhevsky2012imagenet,szegedy2015going,he:15,schroff:15} and image segmentation \cite{long2015fully}. More recently,
they have also shown uncanny abilities to generate images \cite{radford:15,dosovitskiy:14,gregor2015draw}. In particular style transfer \cite{gatys:15,Li2016Combining}, deep dream \cite{mordvintsev:15},
generative adversarial networks \cite{radford:15}, all have been producing highly compelling results. In this work we explore our ability to control the images deep neural networks produce.

For the purposes of this work we use FaceNet \cite{schroff:15}, a face-recognition network that has been trained to distinguish between people, as our
test bench. We address the problem of inverting the network output, or the {\em embedding vector}, i.e., provided with the embedding vector $\Emb$, we generate a
realistic face image, which after being passed through the FaceNet produces $\Emb$.
One interesting aspect of this problem is the fact the space of distinct acceptable solutions is huge, in particular
different orientations and poses of the same person should in theory produce the same embedding.
Furthermore, that space by itself is dominated by the space of
{\it unacceptable} solutions -- the images with glimpses of faces in various orientations, or simply random-noise \cite{szegedy:13}
looking solutions.
All of these unacceptable images are mapped into a given embedding and are thus proper inversions, just not particularly interesting ones.
One approach to solve this is to employ adversarial learning algorithms \cite{radford:15, goodfellow2014generative} where a pair of networks e.g. generator and classifier
are training in parallel. However this somewhat limits our ability to control what is produced by generator.
Our goal for this work is to produce consistent inverse solutions that look like faces in the prescribed position and orientation.
In this paper, perhaps somewhat surprisingly, we show that several very simple regularization techniques worked well in enforcing the consistency of the output images.
In the rest of the section we provide an overview of our results.

\subsection{Image Embedding Network}

  For our experiments we use a Facenet model \cite{schroff:15} mapping a $224\times 224$ RGB face image to a normalized
  128-dimensional ``embedding'' vector. This network was trained to have embeddings of different photographs of the same
  person to be closer to each other than to those of a different person. This network achieves comparable to human-level of face recognition
  performance \cite{schroff:15}.

  \subsection{Overview of results and paper structure}

  The contents of this paper can be roughly separated into two parts.
  First, in \Secs{sec:minimization} and \ref{sec:iterative}, we introduce a general problem of face reconstruction and propose a loss function, using which a gradient-descent style algorithm can reconstruct highly recognizable faces using only the target embedding vector.
  The orientation and facial expression of the produced image match that of a provided guiding image.
  \omtnips{
  The main idea of the method is based on attaching additional regularization losses that enforce face consistency and orientation to the optimized embedding loss function.
  More specifically, we use total-variation loss \cite{mahendran:15} and Laplacian pyramid graident normalization \cite{burt:83}
  to ensure the image is smooth.
  We also use $\ell_2$ distance on intermediate layers with the guiding image to enforce a specific face orientation and position.
  The minimization of the combined loss function is approached by using gradient descent starting at random noise or an apriori chosen initial state.
  }

  In the second part, as outlined in \Sec{sec:ff}, we introduce a feed-forward neural network, which can be trained to produce face images that minimize the loss function used previously for iterative reconstruction.
  We believe this approach to be of independent interest since it allows one to solve the minimization problem in a single step.

  Finally, our experimental results are presented in \Sec{sec:experiments}.
  An interesting observation made while studying the reconstructions, which might be of independent interest, is that
  even faces that look remarkably similar, can still be recognized despite sharing virtually identical macro charactertics.
  We show several examples of this phenomenon in \Sec{sec:experiments}.

\section{Face reconstruction as a minimization problem}
\label{sec:minimization}

  The face reconstruction problem discussed in \Sec{sec:introduction} can be formalized as follows.
  Let $\NetF$ be a function defined by a trained deep neural network, mapping a photo $\Img_\Pers$ of a person $\Pers$ to a lower-dimensional embedding $\Emb_\Pers = \F(\Img_\Pers)$.
  In the following, considering FaceNet, we use two definitions of the embedding vector: an unnormalized embedding obtained in an intermediate FaceNet node and the normalized embedding calulated by applying a softmax activation function to the unnormalized vector.

  Naively, given an embedding $\Emb\in \EmbSpace$, the reconstruction could be accomplished by finding an image $\Img$ minimizing $\Metric\left[\NetF(\Img),\Emb\right]$, where $\Metric:\EmbSpace\times\EmbSpace\to \R$ is some metric on the embedding space $\EmbSpace$.
  However, since in practical applications the space of all possible images $\ImgSpace$ has a much greater dimension that the embedding space $\EmbSpace$, the inverse $\F^{-1}(\Emb)$ of arbitrary $\Emb\in \EmbSpace$ is generally a high-dimensional manifold containing a rich variety of images only a small fraction of which could be considered realisitic face images.
  A more sensible definition of the face reconstruction problem could thus be written as:
  \begin{equation}
       \label{eq:minp}
       \ImgOpt = \argmin_{\Img\in \Faces} \Metric\left[\NetF(\Img),\Emb\right]
       = \argmin_{\Img\in \ImgSpace} \mc{L},
  \end{equation}
  where $\mc{L} = \Metric\left[\NetF(\Img),\Emb\right] + \Reg(\Img)$ and the regularizer $\Reg(\Img)$ vanishes for all images within a subset of ``realistic'' face images $\Faces \subset \ImgSpace$ and $\Reg(\Img)=\infty$ otherwise.
  Since the set $\Faces$ is generally very difficult to define, we solve the minimization problem~\eqref{eq:minp} for other classes of regularization functions $\Reg(\Img)$ which only ``favour'' face-like images.

  One of the approaches to characterizing the set $\Faces$ is based on using a single reference, or a ``guiding'' image $\GImg$.
  Since the trained convolutional neural network defining $\NetF$ contains lower-level ``edge'' and ``shape'' filters as well as more complex features relevant for face recognition, the guiding image regularization function $\GReg(\Img;\GImg)$ can be constructed using the intermediate nodes of this network.
  For example, $\GReg$ could naturally be chosen as $\wG \|\Act(\Img)-\Act(\GImg)\|_r$, where $\wG$ is the regularizer weight and $\Act$ is a vector of activations in a specific network layer $\ell$.
  When $\ell$ is chosen amongst the lowest network layers, the regularizer $\GReg$ effectively pulls $\ImgOpt$ towards the image $\GImg$.
  For higher layers $\ell$, the regularizer $\GReg$ introduces restrictions on the higher-order features of $\ImgOpt$ without necesserily forcing specific textures or colors \cite{gatys:15}.

  The advantage of using a single image $\GImg$ to condition the reconstruction is the possibility to enforce a specific pose, facial expression and background.
  The disadvantage, is of course, the fact that the final image may contain  facial features corresponding to both the embedding and the guiding image.
  For very low values of the guiding image regularizer weight $\wG$, the produced image does not look realistic and frequently consists of numerous face fragments.
  In contrast, for large $\wG$, the reconstruction may be almost indistinguishable from the guiding image.
  By tuning the value $\wG$, it is, however, possible to produce realistic looking faces with barely any facial features ``leaked'' from the guiding image (see \Sec{sec:iterative}).
  \omtnips{
  In Appendix~\ref{sec:gaussian}, we also discuss an alternative approach, in which the regularizer uses a collection of images (with faces sharing a common pose) instead of a single guiding image.
  This regularizer does not force any specific facial features, but generally results in lower-quality images.}

  Numerical optimization of \Eq{eq:minp} frequently produces noisy and distorted images.
  This problem can be alleviated by introducing additional regularizers.
  We use the {\em total-variation} (TV) regularizer \cite{mahendran:15}:
  \begin{equation}
    \label{eq:rtv}
    \RegTV(\Img) = \left[ \| \Img - \ShiftOp_x \Img \|_2^2 + \| \Img - \ShiftOp_y \Img \|_2^2 \right]^{\alpha/2},
  \end{equation}
  which can be seen to penalize images with high-frequency noise, large gradients and sharp boundaries.
  Here, $\ShiftOp_x$ and $\ShiftOp_y$ are operators shifting the entire image by $1$ pixel in $x$ or $y$ direction correspondingly and $\alpha$ is a constant parameter.

  The choice of the optimization function $\Metric$ can have a strong impact on the produced face reconstructions.
  In this paper, we consider two families of loss functions defined on normalized or unnormalized embedding spaces.
  The first one is based on $\ell_2$ metric in the embedding space, \ie $\Metric\left[\NetF(\Img),\Emb\right]=\|\NetF(\Img)-\Emb\|_2^2$.
  Another approach, which was shown to frequently result in higher-quality images, employs a dot-product: $\Metric\left[\NetF(\Img),\Emb\right] = - \NetF(\Img) \cdot \Emb$.

\section{Iterative face reconstruction}
\label{sec:iterative}

  Provided with an embedding $\Emb\in\EmbSpace$ and a chosen set of regularizer parameters, the minimization problem \eqref{eq:minp} can
  be solved numerically using stochastic gradient descent (SGD), Adam \cite{kingma:14},
  or another optimization method starting from a random noise or the guiding image entering $\GReg$.

\begin{figure}
  \centering
  \captionsetup[subfigure]{labelformat=empty}
  \captionsetup[subfigure]{labelformat=parens}
    \begin{subfigure}{.12\textwidth}
        \centering
        \label{fig:guiding}
        \includegraphics[width=.9\linewidth]{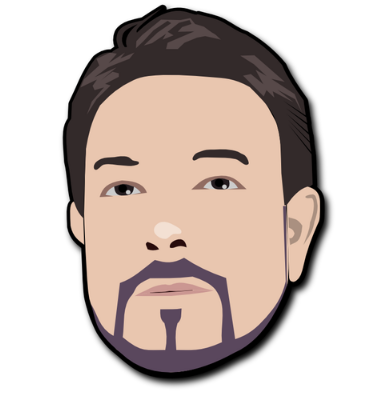}
        \caption{} 
    \end{subfigure}
    \begin{subfigure}{.14\textwidth}
      \centering
      \includegraphics[width=.9\linewidth]{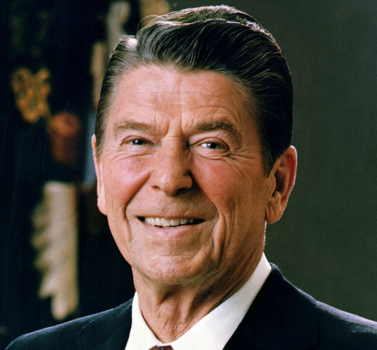}
      \includegraphics[width=.9\linewidth]{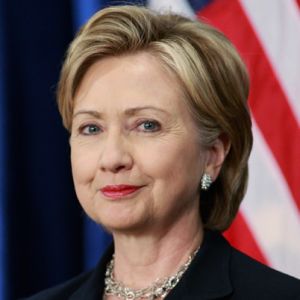}
      \caption{} 
      \label{fig:emb1}
    \end{subfigure}
  \captionsetup[subfigure]{labelformat=parens}
  \begin{subfigure}{.14\textwidth}
      \centering
      \includegraphics[width=.9\linewidth]{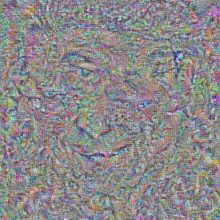}
      \includegraphics[width=.9\linewidth]{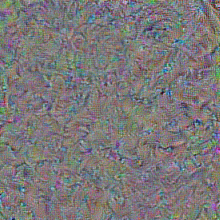}
      \caption{} 
      \label{fig:noguide_no_tv}
    \end{subfigure}
    \begin{subfigure}{.14\textwidth}
      \centering
      \includegraphics[width=.9\linewidth]{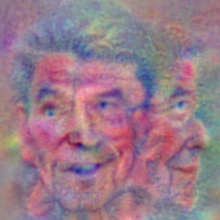}
      \includegraphics[width=.9\linewidth]{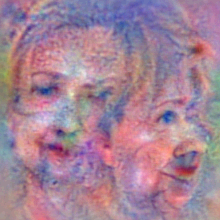}
      \caption{} 
      \label{fig:no_guide_tv}
    \end{subfigure}
    \begin{subfigure}{.14\textwidth}
      \centering
      \includegraphics[width=.9\linewidth]{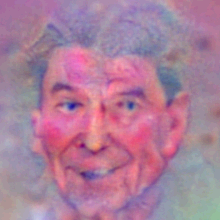}
      \includegraphics[width=.9\linewidth]{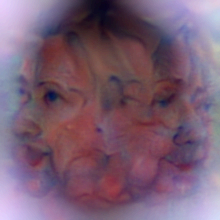}
      \caption{} 
      \label{fig:rand}
    \end{subfigure}
    \begin{subfigure}{.14\textwidth}
      \centering
      \includegraphics[width=.9\linewidth]{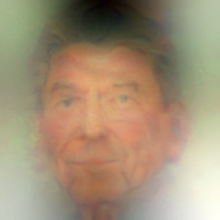}
      \includegraphics[width=.9\linewidth]{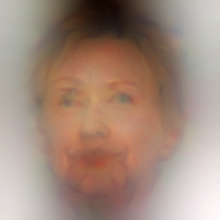}
      \caption{} 
      \label{fig:l2}
    \end{subfigure}
    \begin{subfigure}{.14\textwidth}
      \centering
      \includegraphics[width=.9\linewidth]{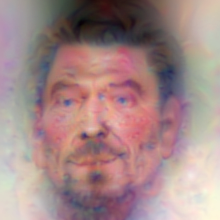}
      \includegraphics[width=.9\linewidth]{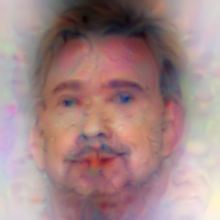}
      \caption{} 
      \label{fig:dot}
    \end{subfigure}
    \caption{Original images and face reconstructions obtained using different techniques:
      (a) Guiding image;
      (b) Source of the embedding used for reconstruction;
      (c) Minimizes $\Metric[\Emb_1,\Emb_2] = \| \Emb_1 - \Emb_2 \|_2^2$ starting from random noise;
      (d) Total Variation added for regularization;
      (e) Total Variation and guiding image regularization added on an intermediate layer;
      (f) Same metric as (e), but with the guiding image initialization;
      (g) Same as (f), but with $\Metric\left[\Emb_1,\Emb_2\right] = - \Emb_1 \cdot \Emb_2$.
    }
  \label{fig:set1}
  \end{figure}

  Without any regularizers, the iterative process converges to an image from within a small neighborhood of the initial state \cite{mahendran:15,szegedy:13}.
  Performing an optimization with the guiding image regularization alone was also unsuccessful at reconstructing a realistic face image.
  A significant improvement was observed once the total-variation regularizer \eqref{eq:rtv} was introduced in \Eq{eq:minp} (see \Figs{fig:l2}, \ref{fig:dot}).
  The initial state of the reconstruction was also shown to play an important role: starting with the guiding image instead of a random noise frequently improved both the
  stability and the quality of the produced images (see \Figs{fig:rand}, \ref{fig:l2}).

  Interestingly, using a sufficiently high $\wG$ allowed us to generate realistic images with facial features of the embedding $\Emb$ and the facial expression of the guiding image.
  By running the algorithm on a sequence of video frames, we were able to perform a ``face transfer'' from the embedding onto the face shown in the video.
  This result is particularly impressive given that the embedding can be produced from just a single photo of a person.

  The positive effect of the TV regularizer has been previously observed, for example, in Ref.~\cite{mahendran:15}.
  One can speculate that it can be attributed to suppresion of high-frequency harmonics leading to the search for a local minimum in a subspace of smooth images.
  \omtnips{
  Indeed, $\pld{\RegTV}{\Img}$ can be shown to be proportional to $\Laplace \Img$, where $\Laplace$ is the discrete Laplacian operator defined as $(\Laplace p)_{x,y} = p_{x+1,y} + p_{x-1,y} + p_{x,y+1} + p_{x,y-1} - 4 p_{x,y}$.
  If the gradient descent step size is sufficiently small, the expression for the SGD update $\Delta \Img = -\mu \, \pld{\mc{L}}{\Img}$, can be viewed as a discretization of a dynamical system with a continuous step number $n$, where the regularizer plays a role of the diffusion term:
  \begin{equation}
    \label{eq:cont}
    \pd{\Img}{n} = -\mu \left( \pd{L}{\Img} - \alpha \wtv \RegTV^{1-\frac{2}{\alpha}} \Laplace \Img \right).
  \end{equation}
  Here $L(\Img)$ is the sum of embedding and guiding image loss functions, which we ultimately need to minimize.
  }
  \omtnips{
  Notice that as shown in \App{sec:minima}, in the limit $\wtv\to \infty$, there are generally a finite number of local minima of $\mc{L}$ and they no longer form high-dimensional submanifolds of $\ImgSpace$.}

  The total-variation regularizer proved to be essential essential for converging towards a realistic face image, however, it also blurs the output images.
  We reduced the blurring effect in the produced reconstruction by reducing $\wtv$ with the iteration number.
  Further we also used laplacian pyramid normalization~\cite{mordvintsev:15} applied to intermediate gradients.
  This improved overall contrast of the image. 

\begin{figure}[b]
  \centering
    \begin{subfigure}{.19\textwidth}
      \centering
      \includegraphics[width=.9\linewidth]{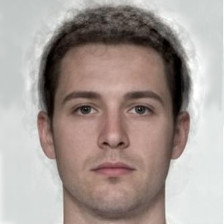}
      \caption{}
      \label{fig:guide1}
    \end{subfigure}
    \begin{subfigure}{.19\textwidth}
      \centering
      \includegraphics[width=.9\linewidth]{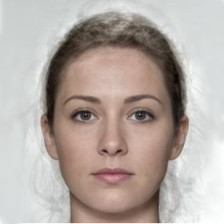}
      \caption{}
      \label{fig:guide2}
    \end{subfigure}
    \begin{subfigure}{.19\textwidth}
      \centering
      \includegraphics[width=.9\linewidth]{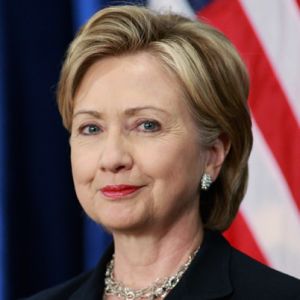}
      \caption{}
      \label{fig:emb4}
    \end{subfigure}
    \begin{subfigure}{.19\textwidth}
      \centering
      \includegraphics[width=.9\linewidth]{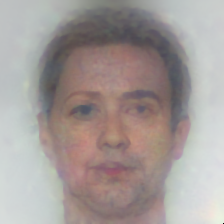}
      \caption{}
      \label{fig:femmale}
    \end{subfigure}
    \begin{subfigure}{.19\textwidth}
      \centering
      \includegraphics[width=.9\linewidth]{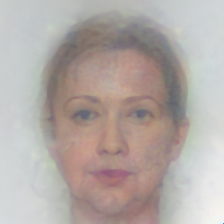}
      \caption{}
      \label{fig:femfem}
    \end{subfigure}
    \caption{Original images and face reconstructions:
      (a) generic ``male'' guiding image;
      (b) generic ``female'' guiding image;
      (c) image used for calculating the target embedding $\Emb$;
      (d) reconstruction of $e$ with the guiding image (a);
      (e) reconstruction of $e$ with the guiding image (b).}
  \label{fig:set2}
  \end{figure}

  The choice of the guiding image had also proven to be very important for a high quality face reconstruction.
  When the guiding image and the embedding corresponded to people of different gender or nationality, the produced images could resemble the guiding image with only some facial features ``borrowed'' from the embedding (see \Figs{fig:femmale}, \ref{fig:femfem}).
  This effect was less noticeable for very large values of $\wtv$, but in this case, the reconstructions had worse quality and were unstable, \ie could be drastically different for different random initial conditions (sometimes producing images with percievably wrong gender or age).
  The problem of the guiding image choice can be solved by either building a classifier which predicts the gender and nationality of the face corresponding to a given embedding (and thus prediciting a proper guiding image to be used in the algorithm), or attempting to produce reconstructions with different guiding images and choosing the result with the smallest embedding space distance.

\section{Feed-forward network for face reconstruction}
\label{sec:ff}

    Each reconstruction obtained using the methods discussed in \Sec{sec:iterative}, requires hundreds or even thousands of iterations.
    Training a feed-forward neural network capable of reconstructing an image in a single pass could have a significant performance advantage.
    The main idea behind training such a network is based on using the same loss function, which we employed for the iterative face reconstruction.
    Specifically, on each training step, given a random input embedding $\Emb$, the network weights are updated to minimize the loss \eqref{eq:minp} calculated on $\Emb$ and the image produced by the network.

\subsection{Feed-forward network taking embedding as an input}
\label{sec:simple-ff}

  The feed-forward network we used for face reconstruction took a $128$-dimensional vector $\Emb$ as an input and produced a $224 \times 224$ image with 3 channels (see \Fig{fig:arch1}).
  Within the network, a fully-connected layer (followed by a ReLU nonlinearity) was first used to transform the embedding vector into a $8\times 8 \times 16$ tensor (with 16 being the number of ``filters'').
  This tensor had then been passed through a sequence of ReLU deconvolutions, each of which took an input tensor of size $2^n \times 2^n \times L_n$ and producing a tensor of the size $2^{n+1} \times 2^{n+1} \times L_{n+1}$.
  In our experiments, the deconvolution kernel had a size $5\times 5$ and all $L_n$ except for the last one with $L_8=3$ were equal.
  The final $256\times 256\times 3$ tensor had been cropped to the fit the FaceNet dimensions of $224\times 224$.
  Using the same loss function which we used for iterative reconstruction, we could then train such a network to produce a face-like image with a desired FaceNet embedding.

\subsection{Feed-forward network with an embedding and a guiding image as inputs}
\label{sec:image-ff}

  A feed-forward network described in \Sec{sec:simple-ff} can be trained to perform face reconstruction with a set of guiding images instead of one.
  In this case, a sparse guiding image index can be passed to the network as one of its inputs.
  Unfortunately, due to a finite capacity of the network and a need to somehow encode all guiding images in the network weights, this could only be demonstrated for a few, but not even dozens of similar guiding images (taken from frames of a video clip).
  One of the approaches to alleviating this restriction and potentially performing face reconstruction with an arbitrary guiding image is based on passing the guiding image as one of the inputs to the feed-forward neural network.

  \begin{figure}
  \centering
  \captionsetup[subfigure]{labelformat=parens}
    \begin{subfigure}{.49\textwidth}
      \includegraphics[width=0.95\linewidth]{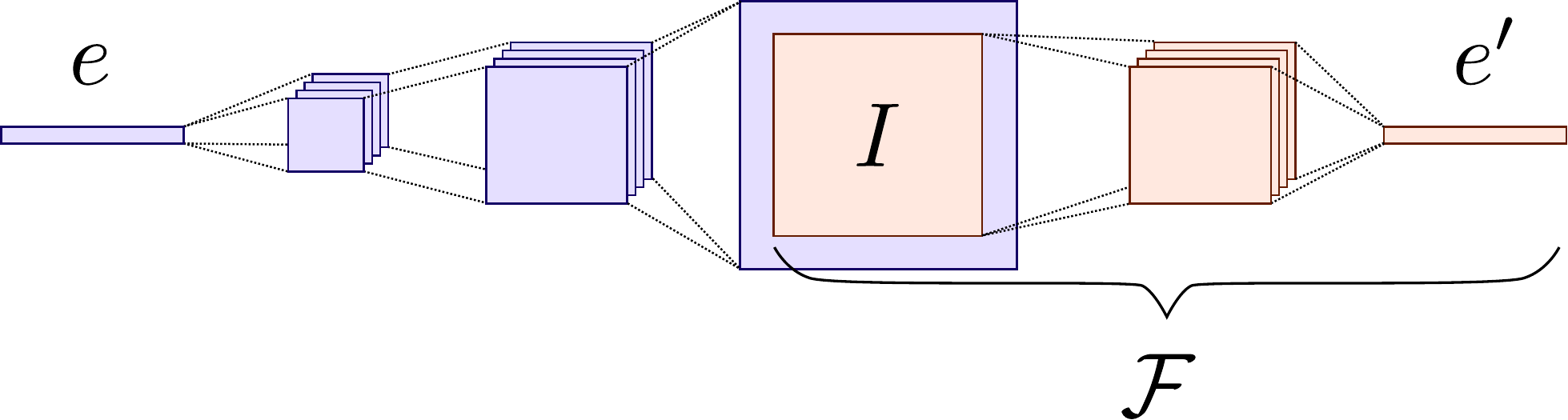}
      \centering
      \caption{}
      \label{fig:arch1}
    \end{subfigure}
    \begin{subfigure}{.49\textwidth}
      \includegraphics[width=0.95\linewidth]{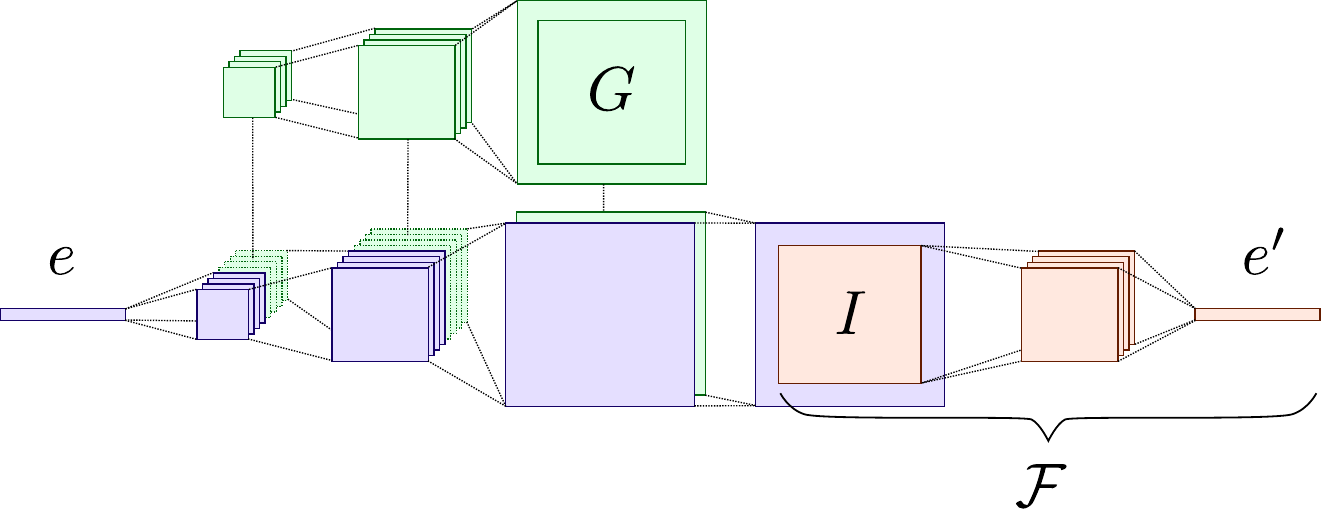}
      \centering
      \caption{}
      \label{fig:arch2}
    \end{subfigure}
    \caption{
      (a) The architecture of a feed-forward network taking an embedding $\Emb$ as an input and producing an output image $I$ (blue).
          The loss function used to train the weights of this network depended on $\Emb'$ obtained by passing the output image $I$ through the FaceNet model ($\F$; red).
      (b) The architecture of a feed-forward network taking an embedding $\Emb$ and the guiding image $G$ as inputs.
          The intermediate tensors obtained by applying a sequence of convolutions with stride 2 to $G$ (green) are concatenated to the intermediate tensors in the feed-forward network (blue).
          The output image $I$ is passed through the FaceNet (red) in order to produce $\Emb'$ entering the loss function.
    }
  \end{figure}

  In our experiments, the input guiding image was first padded to the size of $256\times 256$ and then passed through series of convolutional layers of stride $2$ (see \Fig{fig:arch2}).
  Obtained tensors of size $2^n \times 2^n \times \bar{L}_n$ were then depth-concatenated with the tensors of the same spatial dimensions produced through series of deconvolutions as described above.
  In other words, starting with a $8\times 8\times (16 + \bar{L}_3)$ tensor generated from the embedding vector and the final convolution of the guiding image, each deconvolution consumed a $2^n \times 2^n \times (L_n + \bar{L}_n)$ tensor and produced a $2^{n+1} \times 2^{n+1} \times L_{n+1}$ tensor, which was then concatenated with a $2^{n+1} \times 2^{n+1} \times \bar{L}_{n+1}$ tensor obtained from the guiding image.
  The last $256\times 256 \times (L_8 + 3)$ tensor was finally transformed by a convolution operation to produce an output $256\times 256\times 3$ image.

\section{Experiments}
\label{sec:experiments}

  In this section we explore our ability to generate face images using iterative reconstruction with SGD and feed-forward networks.
  The quality of reconstruction measures the quality of our loss function in finding recognizable faces.
  Once we are satisfied with the loss function, that is: we are reasonably confident that the gradient
  descent with such a loss function produces recognizable faces, we turn our attention to feed-forward networks, which are trained to find the optimum of the same loss, but do it in a single pass.

  For our experiments where we want to illustrate the recognizability of people, we choose to use famous people in order to maximize recognizability of the reconstructed images.
  For our guiding images we use publicly available cartoon images from \Ref{public-domain-vectors} as well as averaged images from \Ref{face-research-org}.

  \subsection{Iterative Reconstruction}

\begin{figure}
      \centering
      \includegraphics[width=0.9\linewidth]{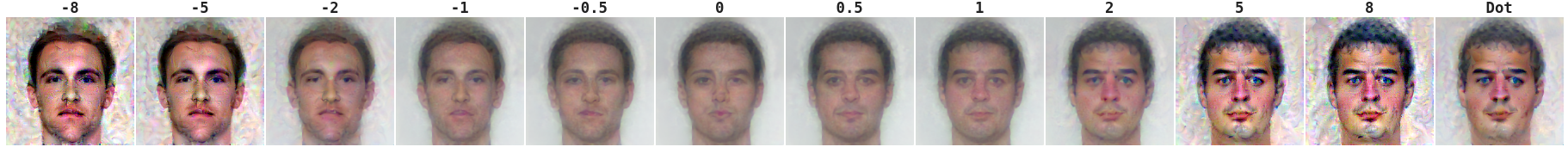}
      \caption{Impact on reconstruction using $\ell_2$ as target metric when the target embedding has been scaled by a given factor.
      Notably, the result obtained with $\ell_2$ metric when the target is scaled is very similar to the result obtained with the dot-product loss (the rightmost image).}
      \label{fig:scaling}
  \end{figure}

  For our experiments we use pre-normalized embedding as an input. Even though, the original FaceNet was trained to differentiate between normalized embeddings and thus ignore
  the difference in $\ell_2$ norm, we find that optimizing a match to pre-normalized embedding produces better results. We conjecture that with normalization, SGD favors smaller embedding values, which
  essentially results in more generic looking image, as illustrated in \Fig{fig:scaling}.
  For our experiments, we use both $\ell_2$ and dot product. Dot product produces significantly sharper, but slightly less
  recognizable images, as demonstrated in \Figs{fig:set1}. Somewhat surprisingly, using normalized $\ell_2$ distance (or, equilvalently, normalized dot product),
  results in much worse reconstructions.
  \begin{figure}
  \centering
  \captionsetup[subfigure]{labelformat=parens}
    \begin{subfigure}{.16\textwidth}
        \centering
        \includegraphics[width=.9\linewidth]{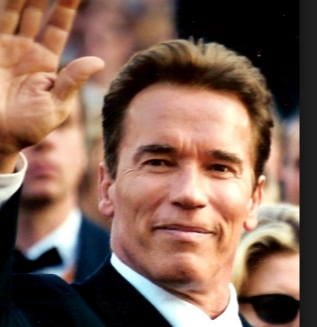}
        \includegraphics[width=.9\linewidth]{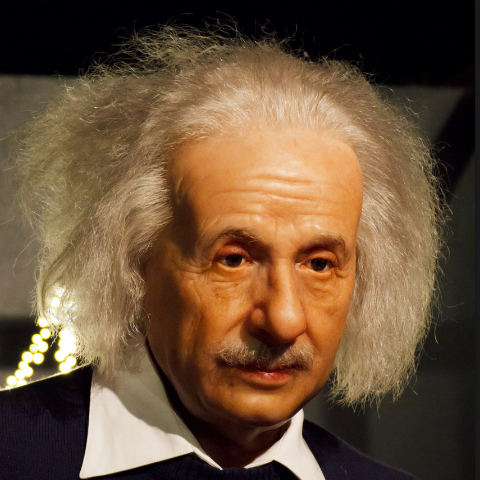}
        \caption{}
    \end{subfigure}
    \begin{subfigure}{.16\textwidth}
      \centering
        \includegraphics[width=.9\linewidth]{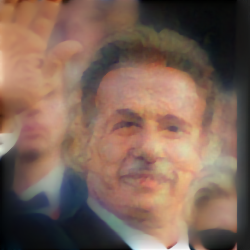}
        \includegraphics[width=.9\linewidth]{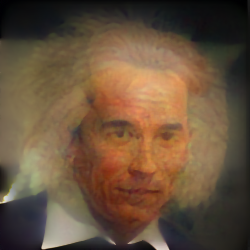}
      \caption{}
      \label{fig:arn_x_ein}
   \end{subfigure}
   \begin{subfigure}{.16\textwidth}
        \centering
        \includegraphics[width=.9\linewidth]{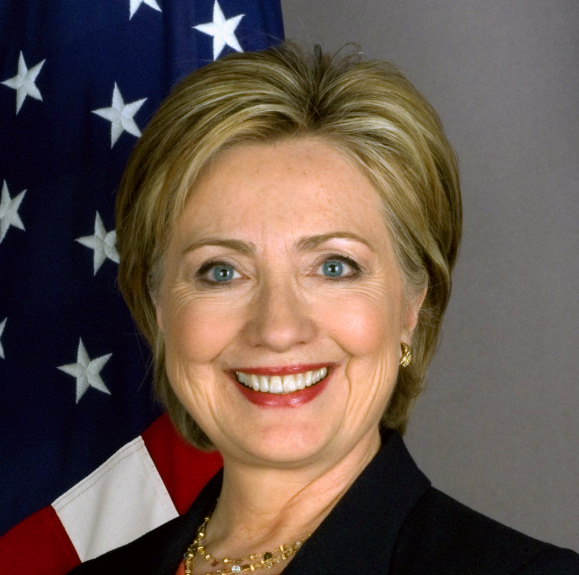}
        \includegraphics[width=.9\linewidth]{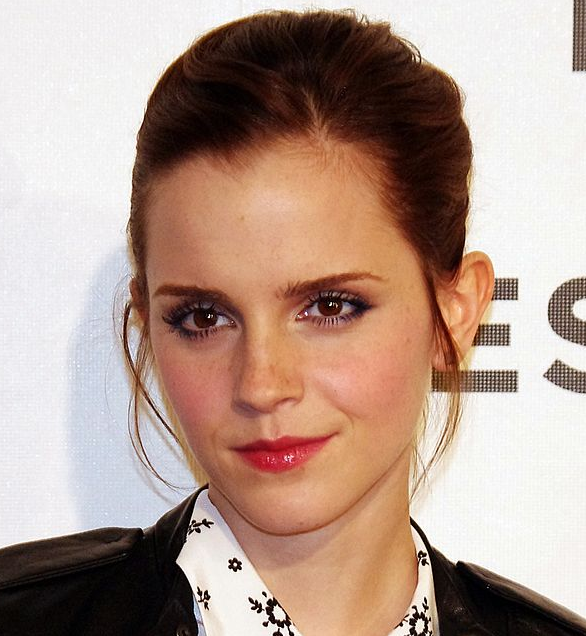}
        \caption{}
    \end{subfigure}
    \begin{subfigure}{.16\textwidth}
      \centering
        \includegraphics[width=.9\linewidth]{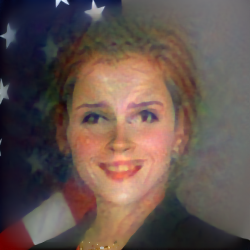}
        \includegraphics[width=.9\linewidth]{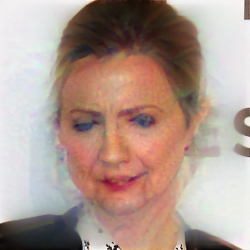}
      \caption{}
      \label{fig:watson_x_clinton}
   \end{subfigure}
  \caption{Face transfer: An embedding of one person transferred over to the photograph of another. In all cases network used one photo as guide, and embedding of the other as target.}
  \end{figure}

  \begin{figure}
  \centering
    \includegraphics[width=.85\linewidth]{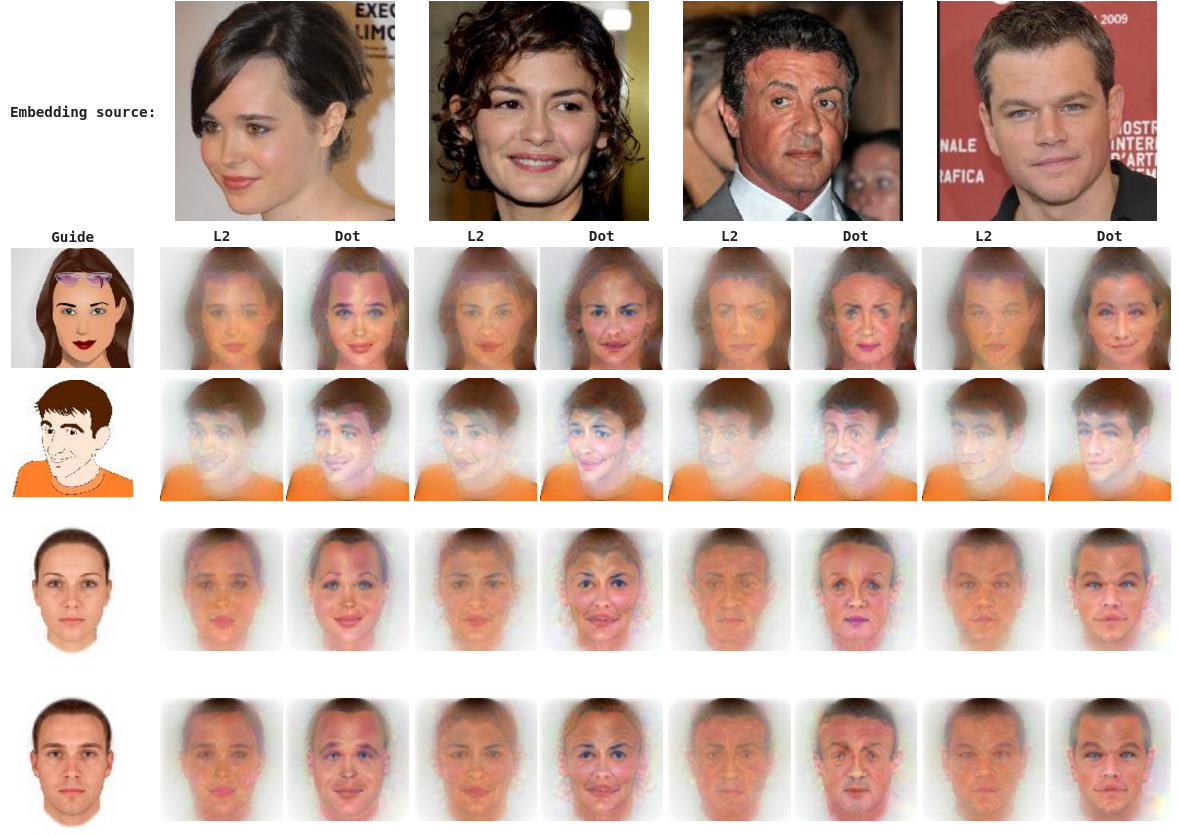}
    \caption{Face transfer examples. It is noteworthy that different people have different degree of recognizability, and some are strong enough to change the
    gender of the guiding image, and some are not.}
    \label{fig:manypeople}
  \end{figure}

  For the example in \Fig{fig:set1} we have used activation target on a single intermediate layer.
  But in the remainder of this section, we attach $\ell_2$-distance loss to multiple intermediate layers as well as use
  Laplacian Pyramid Gradient Normalization~\cite{mordvintsev:15} to change spectral characteristics of the image gradients.
  We find that this technique improves the quality of reconstructed images.
  In figure \Fig{fig:manypeople} we show face reconstructions for multiple
  celebrities. The guiding images that we use are either cartoon-like faces, or average faces from \Ref{face-research-org}. For more face transfer examples including face transfer in a video see supplementary materials.

\begin{figure}
      \centering
      \begin{subfigure}{.68\textwidth}
        \includegraphics[width=0.9\linewidth]{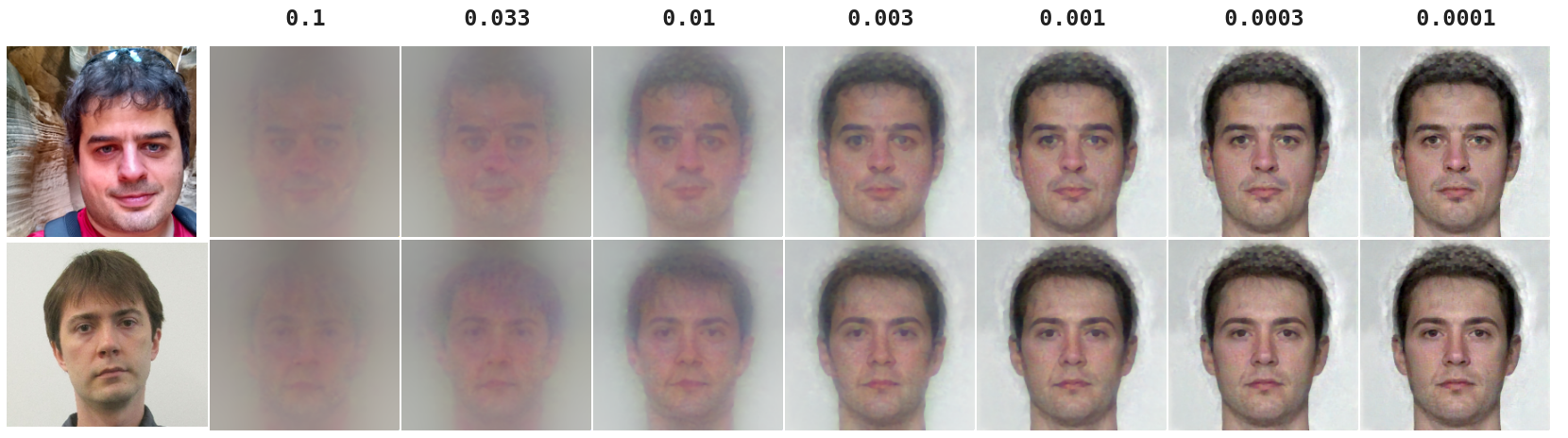}
        \caption{}
      \end{subfigure}
      \begin{subfigure}{.3\textwidth}
      \centering
      \includegraphics[width=0.99\linewidth]{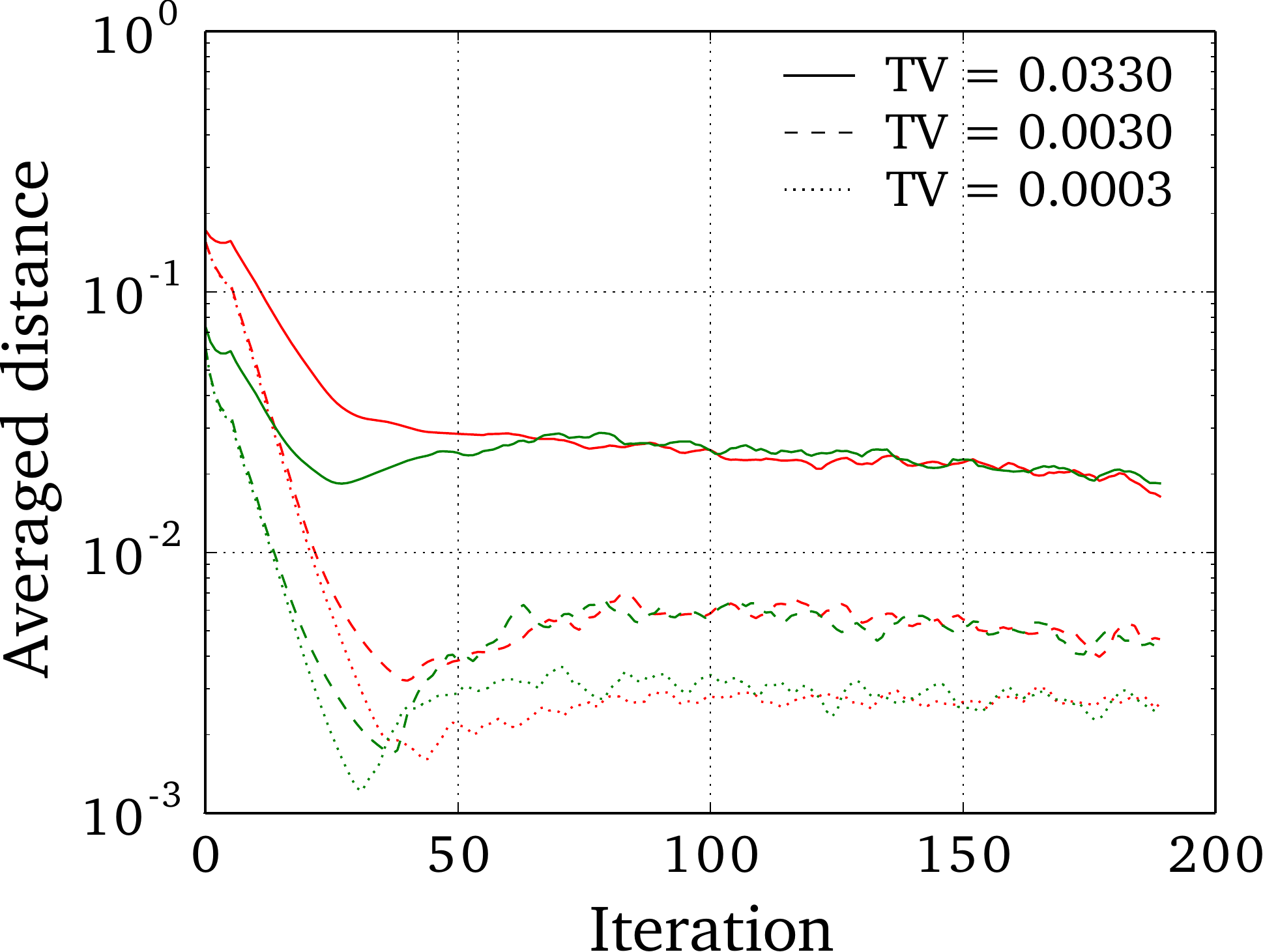}
      \caption{}
      \label{fig:loss_for_high_tv}
    \end{subfigure}
      \caption{
      The impact of changing the TV weight. The first column contains the original images, whose embeddings we used.
      To highlight the difference we did not apply brightness correction. \omtnips{On figure \Fig{fig:normalized} we show the first three images with adjusted brighness and intensity.}
      (b) Distance to the embedding for different values as a function of gradient descent iteration.
      Note: more recognizable images are further away from the target embedding.}
      \label{fig:impact_of_tv}
  \end{figure}

  We then turn our attention to the total-variation regularizer. In figures \Fig{fig:impact_of_tv}, we show how increasing TV weight affects the image quality. To demonstrate the impact of high TV we did not perform
  any normalization of the image, which results in very subdued images.
  It is interesting to note that
  the embedding $\Emb^*$ of the reconstructed image gets further away from the target as shown in \Fig{fig:loss_for_high_tv} and yet the
  image becomes more recognizable. Another observation that might be of independent interest
  is that images in the right-most column of \Fig{fig:impact_of_tv} (the lowest total variation), are extremely similar, and yet each exhibit some traits of the person whose embedding they minimize.

  \omtnips{
  \begin{figure}
      \centering
      \includegraphics[width=0.9\linewidth]{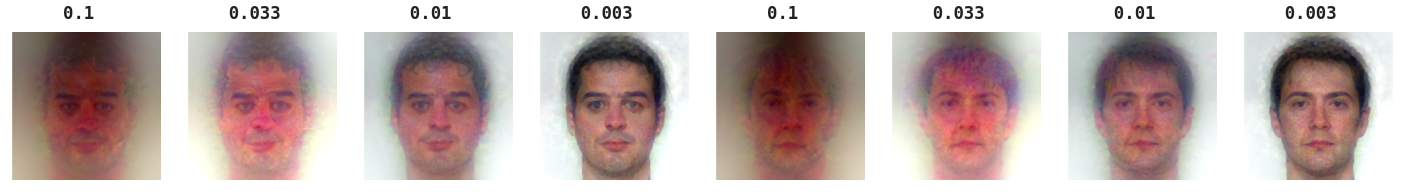}
      \caption{Same as \Fig{fig:impact_of_tv} with adjusted brightness and contrast.}
      \label{fig:normalized}
  \end{figure}
  }

  \subsection{Feed-forward network}

\begin{figure}[b]
    \begin{minipage}{0.6\textwidth}
      \centering
      \begin{subfigure}{.24\textwidth}
        \centering
        \includegraphics[width=.9\linewidth]{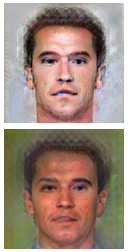}
        \caption{}
      \end{subfigure}
      \begin{subfigure}{.24\textwidth}
        \centering
        \includegraphics[width=.9\linewidth]{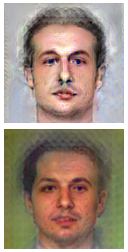}
        \caption{}
      \end{subfigure}
      \begin{subfigure}{.24\textwidth}
        \centering
        \includegraphics[width=.9\linewidth]{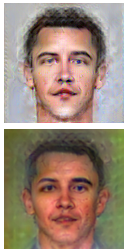}
        \caption{}
      \end{subfigure}
      \begin{subfigure}{.24\textwidth}
        \centering
        \includegraphics[width=.9\linewidth]{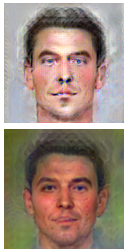}
        \caption{}
      \end{subfigure}
      \captionof{figure}{Face reconstructions obtained using a feed-forward network trained with a generic male image from \Fig{fig:guide1}:
    (a) Arnold Schwarzenegger (b) Albert Einstein; (c) Barack Obama; (d) Ronald Reagan.
     Top row uses a dot-product embedding loss and the bottom row uses $\ell_2$ distance.}
        \label{fig:simple-ff}
    \end{minipage}
   \hspace{0.3cm}
    \begin{minipage}{.4\textwidth}
      \centering
      \includegraphics[width=0.98\linewidth]{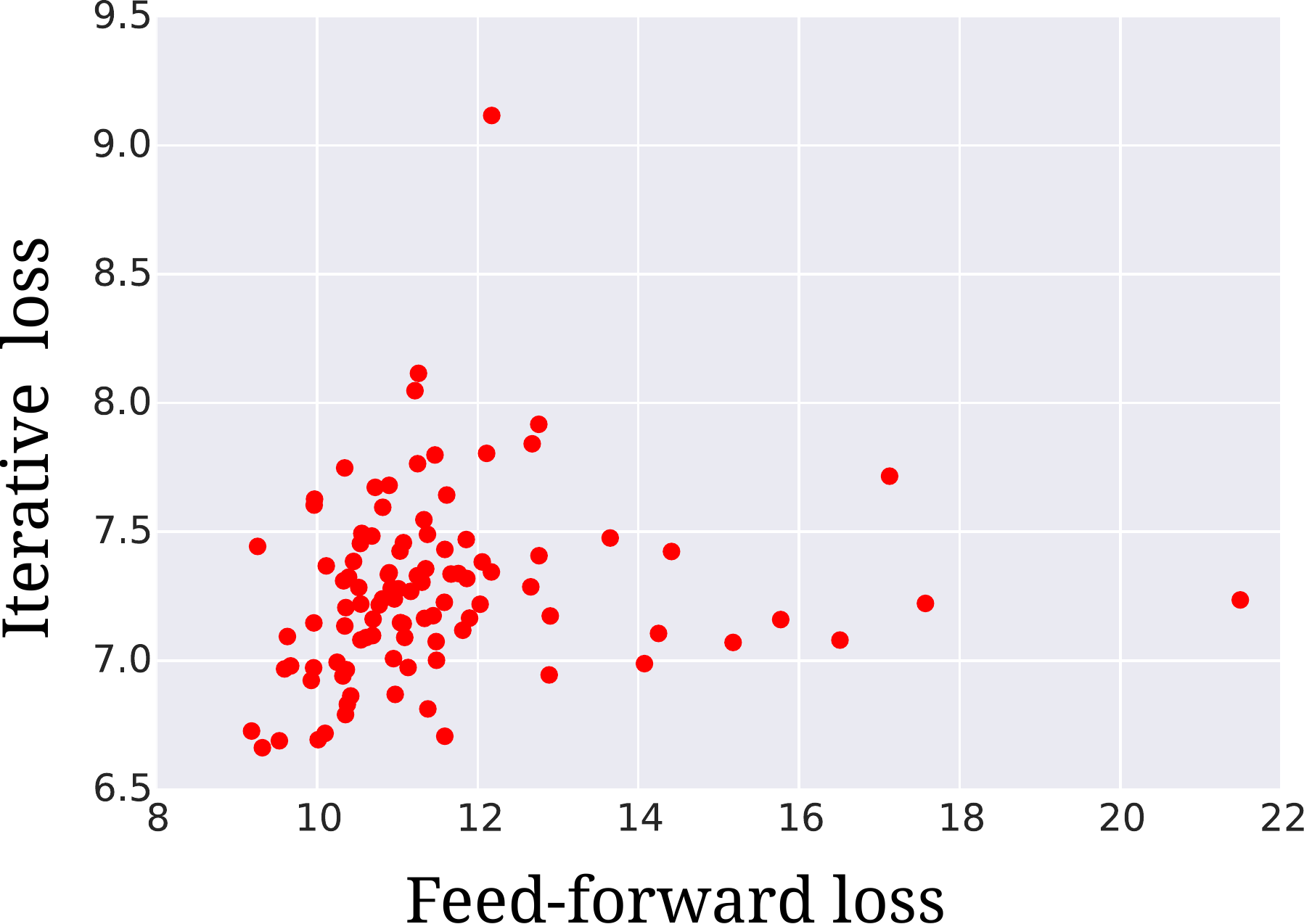}
      \captionof{figure}{Scatter plot obtained for 100 embeddings, showing a feed-forward loss function $\mc{L}$ against  iterative loss achieved by minimizing $\mc{L}$ using SGD.}
      \label{fig:scatter}
    \end{minipage}
  \end{figure}

  Being trained on a set of embeddings and a single guiding image, the feed-forward network described in \Sec{sec:simple-ff} taking a FaceNet embedding as an input and producing an image as its output, succesfully converged.
  The images produced by the network on several embeddings never seen during the training are shown in \Fig{fig:simple-ff}.
  The model was observed to converge faster if the network weights were initialized as follows.
  The deconvolution weights were chosen to produce smooth spatial interpolation with random Xaiver filter-to-filter connections and the final deconvolution was tuned to produce grayscale output image.

  It is remarkable that seemingly similar images produced with the feed-forward network are still recognizible reconstructions of the provided embeddings.
  This is a demonstration of the fact that only a subtle change of facial features is frequently sufficient to distinguish one person from another.

  \omt{For example one could compare the pixel difference between different images.
  Averaged over a set of pairs of different embeddings, this difference could characterize the extent to which the network modifies the guiding image in order to match the input embedding.}

  \omtnips{While in case of iterative reconstruction, there is generally no expectation of the reconstructed image to be smooth, unless we apply a regularizer, for convolutional networks it is
  smoothness is faciliatedby the fact that lots of parameters are shared across the entire image. One approach quantifying the quality of the constructed feed-forward network is to use the
  average distance between reconstructed and target embeddings.}
  Note that in contrast with iterative reconstruction, the feed-forward network is unlikely to match input embeddings almost exactly by means of adding just a small perturbation to the guiding image.
  Indeed, since this additive perturbation is expected to strongly depend on the input embedding, ``memorizing'' this complicated dependence may require much greater network information capacity
  than producing accurate smooth reconstructions. On the other hand, the final embedding space loss calculated for images produced by the feed-forward network can be used for choosing optimal
  model parameters.
  Table~\ref{table:filters} shows average values of the total loss function, $\ell_2$ embedding space distance and the dot product for the normalized embeddings calculated for several trained feed-forward networks (with $\ell_2$ distance optimization) on a set of 100 embedding vectors.
  Even though the feed-forward network seems to perform best with the largest number of filters, using just $50$ filters already results in $\avr{\tilde{e}_1\cdot \tilde{e}_2} \approx 0.75$, which is greater than $0.6$, the average value generally obtained for different real photos of a single person.

  \newcommand\Tstrut{\rule{0pt}{2.6ex}}
  \newcommand\Bstrut{\rule[-0.9ex]{0pt}{0pt}}

  \begin{table}[b]
    \centering
    \begin{tabular}{ c | c | c | c | c | c | c |}
        \cline{2-7}
        &
        \multicolumn{3}{|c|}{$\wtv = 10^{-4}$\Tstrut\Bstrut} &
        \multicolumn{3}{|c|}{$\wtv = 10^{-3}$} \\
        \hline
        \multicolumn{1}{|c|}{Number of filters $(L_n)$\Tstrut\Bstrut} &
        \multicolumn{1}{|c|}{$\avr{\mc{L}}$} &
        \multicolumn{1}{|c|}{$\avr{\|e_1-e_2\|_2^2}$} &
        \multicolumn{1}{|c|}{$\avr{\tilde{e}_1\cdot \tilde{e}_2}$} &
        \multicolumn{1}{|c|}{$\avr{\mc{L}}$} &
        \multicolumn{1}{|c|}{$\avr{\|e_1-e_2\|_2^2}$} &
        \multicolumn{1}{|c|}{$\avr{\tilde{e}_1\cdot \tilde{e}_2}$} \\
        \hline
        \multicolumn{1}{|c|}{50\Tstrut} & 15.098 & 0.0544 & 0.752 & 16.927 & 0.0533 & 0.751 \\
        \multicolumn{1}{|c|}{112} & 12.089 & 0.0407 & 0.804 & 14.798 & 0.0431 & 0.788 \\
        \multicolumn{1}{|c|}{250\Bstrut} & 11.142 & 0.0352 & 0.832 & 11.615 & 0.0400 & 0.808 \\
        \hline
    \end{tabular}
    \vskip .2cm
    \caption{
      Dependence of the average total loss $\mc{L}$, average $\ell_2$ distance $\avr{\|e_1-e_2\|_2^2}$ and average dot product $\avr{\tilde{e}_1\cdot \tilde{e}_2}$ calculated for a
     feed-forward network on the number of filters $L_n$ and the TV weight $\wtv$.
      Here $e_1$,  $e_2$ are the unnormalized embeddings of the target and reconstruction respectively. Tilde denotes normalized value.
    }
    \label{table:filters}
  \end{table}
  The extent to which the trained feed-forward network optimizes the loss function $\mc{L}$ can be compared to that of the iterative reconstruction algorithm.
  Figure~\ref{fig:scatter} shows a scatter plot comparing the values of $\mc{L}$ obtained using two approaches.
  As one would expect, the average of the full loss function calculated for the feed-forward solutions (for $\ell_2$ embedding space loss and a set of 100 random embeddings) is by a factor of $1.6$ greater than the average loss obtained via iterative reconstruction for the same embeddings.
  Interestingly, the results of the iterative reconstructions are percieved to be worse than those produced by the feed-forward network.
  At the same time, the average embedding space distance between the input and output is significantly smaller for the iterative reconstruction.

\begin{figure}[t]
  \centering
    \begin{subfigure}{.14\textwidth}
      \centering
      \includegraphics[width=.9\linewidth]{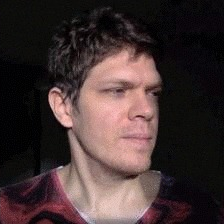}
      \caption{}
    \end{subfigure}
    \begin{subfigure}{.14\textwidth}
      \centering
      \includegraphics[width=.9\linewidth]{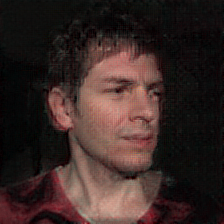}
      \caption{}
    \end{subfigure}
    \begin{subfigure}{.14\textwidth}
      \centering
      \includegraphics[width=.9\linewidth]{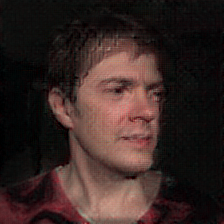}
      \caption{}
    \end{subfigure}
    \begin{subfigure}{.14\textwidth}
      \centering
      \includegraphics[width=.9\linewidth]{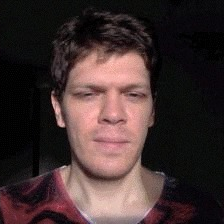}
      \caption{}
    \end{subfigure}
    \begin{subfigure}{.14\textwidth}
      \centering
      \includegraphics[width=.9\linewidth]{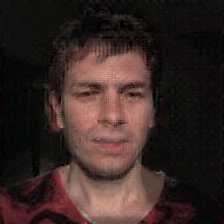}
      \caption{}
    \end{subfigure}
    \begin{subfigure}{.14\textwidth}
      \centering
      \includegraphics[width=.9\linewidth]{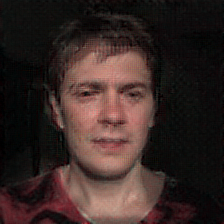}
      \caption{}
    \end{subfigure}
    \caption{Original guiding images [(a) and (d)] and face reconstructions obtained using a feed-forward network that takes the guiding image as one of the inputs. The network has been trained on 70 frames of a short video clip:
    (a), (b) and (c) 1st orientation; (d), (e) and (f) 2nd orientation.}
  \label{fig:image-ff}
  \end{figure}

  \paragraph{Feed-forward network with an embedding and a guiding image as inputs}
  In our experiments, we trained a feed-forward network described in \Sec{sec:image-ff} on random embedding vectors and $70$ frames of a short video clip treated as a set of independent guiding images.
  After the network had been trained, we used an embedding vector not seen during the training stage and the same frame sequence to produce an animation.
  This procedure is generally sucessfull at performing face transfer from the embedding to the target video clip.
  A few frames from the resulting animation are shown in \Fig{fig:image-ff}.

\section{Conclusions and future work}

  We have demonstrated that a FaceNet embedding loss coupled with simple regularization functions can be used to succesfully reconstruct realistic looking faces.
  Both gradient descent and simple deep neural networks were shown to produce high-quality results.
  In a way, our work defines two distinct areas that should guide future
  research.
  On one hand, we can explore better regularization functions that might improve the quality of the generated images
  and combine multiple networks for more precise control of the reconstructions.
  For example, it would be interesting to explore the capability of controlling facial expression of the generated image using a network that was
  trained to recognize facial expressions (such as anger, satisfaction, smile etc.).

  On the other hand, in order to achieve fast generation we need to train a deep network that would solve the minimization problem in
  one pass.
  It is interesting that we were able to achieve this without using adversarial learning.
  This suggests that embedding produced by unrelated network is mostly ``complete'' in the sense
  that it contains enough information to reconstruct a recognizable image that matches the original in human understandable sense. An interesting further extension would be to employ more advanced architectures
  including those utilizing recurrent networks. Also, combining our techniques and adversarial learning is a very promising direction.

\paragraph{Acknowledgments.}
  The authors thank Alexander Mordvintsev, Blaise Ag\"uera y Arcas, Eider Moore and Florian Schroff for valueable discussions.

\small

\bibliographystyle{unsrt}
\bibliography{main.bib}

\begin{thebibliography}{10}

\bibitem{krizhevsky2012imagenet}
Alex Krizhevsky, Ilya Sutskever, and Geoffrey~E Hinton.
\newblock Imagenet classification with deep convolutional neural networks.
\newblock In {\em Advances in neural information processing systems}, pages
  1097--1105, 2012.

\bibitem{szegedy2015going}
Christian Szegedy, Wei Liu, Yangqing Jia, Pierre Sermanet, Scott Reed, Dragomir
  Anguelov, Dumitru Erhan, Vincent Vanhoucke, and Andrew Rabinovich.
\newblock Going deeper with convolutions.
\newblock In {\em Proceedings of the IEEE Conference on Computer Vision and
  Pattern Recognition}, pages 1--9, 2015.

\bibitem{he:15}
Kaiming He, Xiangyu Zhang, Shaoqing Ren, and Jian Sun.
\newblock Deep residual learning for image recognition.
\newblock 2015.
\newblock \href{http://arxiv.org/abs/1512.03385}{arXiv:cs.CV/1512.03385}.

\bibitem{schroff:15}
Florian Schroff, Dmitry Kalenichenko, and James Philbin.
\newblock {FaceNet}: {A} unified embedding for face recognition and clustering.
\newblock In {\em Proceedings of the IEEE Conference on Computer Vision and
  Pattern Recognition}, pages 815--823, 2015.

\bibitem{long2015fully}
Jonathan Long, Evan Shelhamer, and Trevor Darrell.
\newblock Fully convolutional networks for semantic segmentation.
\newblock In {\em Proceedings of the IEEE Conference on Computer Vision and
  Pattern Recognition}, pages 3431--3440, 2015.

\bibitem{radford:15}
Alec Radford, Luke Metz, and Soumith Chintala.
\newblock Unsupervised representation learning with deep convolutional
  generative adversarial networks.
\newblock 2015.
\newblock \href{http://arxiv.org/abs/1511.06434}{arXiv:cs.LG/1511.06434}.

\bibitem{dosovitskiy:14}
Alexey Dosovitskiy, Jost~Tobias Springenberg, and Thomas Brox.
\newblock Learning to generate chairs with convolutional neural networks.
\newblock In {\em IEEE International Conference on Computer Vision and Pattern
  Recognition}, 2014.

\bibitem{gregor2015draw}
Karol Gregor, Ivo Danihelka, Alex Graves, and Daan Wierstra.
\newblock Draw: A recurrent neural network for image generation.
\newblock {\em arXiv preprint arXiv:1502.04623}, 2015.

\bibitem{gatys:15}
Leon~A. Gatys, Alexander~S. Ecker, and Matthias Bethge.
\newblock A neural algorithm of artistic style.
\newblock 2015.
\newblock \href{http://arxiv.org/abs/1508.06576}{arXiv:cs.CV/1508.06576}.

\bibitem{Li2016Combining}
Chuan Li and Michael Wand.
\newblock Combining markov random fields and convolutional neural networks for
  image synthesis.
\newblock {\em CoRR}, abs/1601.04589, 2016.

\bibitem{mordvintsev:15}
A.~Mordvintsev, C.~Olah, and M.~Tyka.
\newblock "deepdream - a code example for visualizing neural networks", 2015.

\bibitem{szegedy:13}
Christian Szegedy, Wojciech Zaremba, Ilya Sutskever, Joan Bruna, Dumitru Erhan,
  Ian Goodfellow, and Rob Fergus.
\newblock Intriguing properties of neural networks.
\newblock 2013.
\newblock \href{http://arxiv.org/abs/1312.6199}{arXiv:cs.CV/1312.6199}.

\bibitem{goodfellow2014generative}
Ian Goodfellow, Jean Pouget-Abadie, Mehdi Mirza, Bing Xu, David Warde-Farley,
  Sherjil Ozair, Aaron Courville, and Yoshua Bengio.
\newblock Generative adversarial nets.
\newblock In {\em Advances in Neural Information Processing Systems}, pages
  2672--2680, 2014.

\bibitem{mahendran:15}
Aravindh Mahendran and Andrea Vedaldi.
\newblock Understanding deep image representations by inverting them.
\newblock In {\em Computer Vision and Pattern Recognition (CVPR), 2015 IEEE
  Conference on}, pages 5188--5196. IEEE, 2015.

\bibitem{burt:83}
Peter~J. Burt and Edward~H. Adelson.
\newblock The laplacian pyramid as a compact image code.
\newblock {\em IEEE Transactions on Communications}, 31(4):532--540, 1983.

\bibitem{kingma:14}
Diederik Kingma and Jimmy Ba.
\newblock Adam: A method for stochastic optimization.
\newblock 2014.
\newblock \href{http://arxiv.org/abs/1412.6980}{arXiv:cs.LG/1412.6980}.

\bibitem{public-domain-vectors}
publicdomainvectors.org.

\bibitem{face-research-org}
Lisa DeBruine and Ben Jones.
\newblock "http://faceresearch.org: Experiments about faces and voice
  preferences".

\bibitem{schaaf:96}
A.~Van der Schaaf and JH~van Hateren.
\newblock Modelling the power spectra of natural images: statistics and
  information.
\newblock {\em Vision research}, 36(17):2759--2770, 1996.

\end{thebibliography}

\normalsize

\omtnips{
  \newpage
  \setcounter{secnumdepth}{0}
  \section{Appendices}
  \setcounter{secnumdepth}{1}
  \appendix

  \section{Gaussian activation regularizer}
  \label{sec:gaussian}

    Instead of using a single image for regularization, one could consider a collection of photos $\SetImgs$, faces in which share some characteristics like position, pose or facial expression.
    Given a function $\Prob_\SetImgs$ modeling a distribution of images $\SetImgs$ and some constant $\epsilon$, the regularizer $\Reg$ could, for example, be defined as $\Reg(\Img) = 0$ for $\Prob_\SetImgs(\Img) > \epsilon$ and $\Reg(\Img)=\infty$ otherwise.
    However, since many numerical optimization methods perform better on smooth functions, a more practically suitable choice of $\Reg$ could be $\RegGauss(\Img) \propto -\mu \log \Prob_\SetImgs(\Img)$ with $\Prob_\SetImgs$ modelled by a Gaussian distribution in the activation space:
    \begin{equation}
        \label{eq:P}
        \Prob_\SetImgs = C \exp\left(
            - \sum_{n\in \ell} \frac{(\Act_n - \Av_n)^2}{2\Dev^2_n}
        \right),
    \end{equation}
    where $C$ is a normalization constant, $n$ goes over all nodes in the layer $\ell$, $\Act_n$ are node activations and $\Av_n$, $\Dev_n$ are the average values and standard deviations of the activations $\Act_n$ computed for all images in $\SetImgs$.
    For lower layers $\ell$, $\RegGauss$ is expected to penalize images with colors or textures inconsistent with those present in the majority of images from $\SetImgs$.
    For higher layers, in turn, $\RegGauss$ would give preference to images with the ``right'' higher-order features.

    Notice that $\RegGauss$ given by \Eq{eq:P} can also be thought as a more ``natural'' way of defining a metric in the activation space.
    Unlike the guiding image regularizer $\GReg$, which arbitrarily uses an $\ell_2$ metric, \Eq{eq:P} is invariant under linear rescaling of individual activations.

    In practical applications, some of the neural network activations may be almost independent of the network input.
    Since the corresponding terms dominate $\RegGauss$, we introduce a small parameter $\nu$ ``smoothing'' the regularization function and preventing singularities:
    \begin{equation}
        \label{eq:rgauss}
        \RegGauss(\Img) = \sum_{n\in \ell} \frac{\nu \left( \Act_n - v_n \right)^2 \sigma_{\ell\,{\rm max}}^2}{\sigma_n^2 + \nu \sigma_{\ell\,{\rm max}}^2},
    \end{equation}
    where $\sigma_{\ell\,{\rm max}} = \max_{n\in \ell} \sigma_n$.
    In our face reconstruction experiments, we used this final form \eqref{eq:rgauss} of the regularizer.

    In contrast to experiments with a single guiding image, the reconstructions produced with this regularizer do not inherit facial features from any specific pre-defined image.
    However, they also tend to be less photo-realistic since the average activations $\{\Av_n\}$ include ``traces'' of numerous images.
    At lower layers, for example, $\{\Av_n\}$ described a blurred image obtained by superimposing all images from the collection $\SetImgs$.

  \section{Local minima of $\mc{L}$ in the limit $\wtv\to \infty$}
  \label{sec:minima}

    The stationary states of \Eq{eq:cont} satisfying
    \begin{equation}
      \label{eq:limg}
      \Laplace \Img = Z^{-1} \pd{L}{\Img},
    \end{equation}
    where $Z(\Img)=\alpha \wtv \RegTV^{1-\frac{2}{\alpha}}(\Img)$, can be found asymptotically as $\wtv\to\infty$.
    Assuming that the shift operators $\ShiftOp_x$ and $\ShiftOp_y$ are cyclic, one can perform a discrete Fourier transformation of $\Img$:
    \begin{equation*}
      \Img_{x,y}=\sum_{n_x=0}^{N-1}\sum_{n_y=0}^{N-1} e^{2\pi i (n_x x + n_y y) / N} \tilde{\Img}_{n_x,n_y}
    \end{equation*}
    and the entire equation~\eqref{eq:limg}:
    \begin{equation}
      \label{eq:sv}
      -4 \gamma_{n_x,n_y} \tilde{\Img}_{n_x,n_y}
      =
      Z^{-1} \left(\pd{L}{\Img}\right)_{n_x,n_y},
    \end{equation}
    where $\gamma_{n_x,n_y} = \sin^2 \left( \pi n_x / N \right) + \sin^2 \left( \pi n_y / N \right)$.
    Since $\tilde{\Img}_{n_x,n_y} = O(Z^{-1})$ for all non-zero $(n_x,n_y)$, we can rewrite \Eq{eq:sv} as:
    \begin{align}
      \label{eq:sv1}
      - 4 \gamma_{n_x,n_y} Z \delta \tilde{\Img}_{n_x,n_y} =&
      \left(\pd{L}{\Img}\right)_{n_x,n_y} (\bar{\Img}) +
      \pd{}{\Img} \left(\pd{L}{\Img}\right)_{n_x,n_y} \delta \Img +
      O(\delta p^2), \\
      \label{eq:sv2}
      0 =&
      \left(\pd{L}{\Img}\right)_{0,0} (\bar{\Img}) + \pd{}{\Img} \left(\pd{L}{\Img}\right)_{0,0} \delta \Img + O(\delta p^2),
    \end{align}
    where the stationary state $\Img$ is expressed as a sum of a constant component $\bar{\Img}=\avr{\Img}$ and $\delta \Img=\Img-\avr{\Img}$.

    Equations~\eqref{eq:sv1} and \eqref{eq:sv2} can be solved approximately by expanding both $\bar{\Img}$ and $\delta \Img$ in the powers of $Z^{-1}$.
    In the lowest order, $\bar{\Img}$ satisfies
    \begin{equation}
      \label{eq:avimg}
      \left(\pd{L}{\Img}\right)_{0,0}(\bar{\Img}) = 0
    \end{equation}
    and $\delta \Img = C\delta \Img'$ with
    \begin{equation}
      \label{eq:dimg}
      \delta \Img' = -\frac{1}{4 \alpha \gamma_{n_x,n_y} \wtv} \left(\pd{L}{\Img}\right)_{n_x,n_y} (\bar{\Img})
    \end{equation}
    and $C = R_{\rm TV}^{\frac{2-\alpha}{\alpha(\alpha-1)}}(\delta p')$.

    Since \Eq{eq:avimg} generally has a finite number of solutions, we expect that there is a finite number of local minima of $\mc{L}$ for $\wtv\to \infty$.
    Furthermore, noticing that $\gamma_{n_x,n_y} \propto n_x^2 + n_y^2$, one can see from \Eq{eq:dimg} that the total-variation regularizer indeed suppresses higher harmonics of $\pld{L}{\Img}$.


  \section{Other approaches for improving image quality}
  \label{sec:regrest}

    Natural images are typically characterized by an intensity power spectrum $I(f_x,f_y)$ obeying \cite{schaaf:96} an approximate power law $I\sim (f_x^2+f_y^2)^{-1}$.
    The face reconstruction algorithm could be regularized to produce images with a tuned spectrum by performing a Laplacian pyramid (LP) decomposition \cite{burt:83} of the image.
    Let $g_0$ be the original image, $\op{e}$ be the ``expand'' operator and $\op{r}$ the ``reduce'' operator \cite{burt:83}.
    The Laplacian pyramid can then be defined as a sequence of images $L_k = g_k - \op{e} g_{k+1}$, where $g_{k+1} = \op{r} g_k$.
    The LP normalization regularizer
    \begin{equation}
      \label{eq:rlp}
      \RegLP(\Img) = \sum_{n=1}^{N} \left( \| \LPComponent_n(p) \| - \LPNorm 2^{\beta n}\right)^2,
    \end{equation}
    can then favour images with the desired power spectrum $\beta$ and a component norm $\LPNorm$.
    An alternative approach is based on normalizing the Laplacian pyramid components of the gradient updates themselves.

    In a case when the reconstruction is expected to respect a particular symmetry, the optimization problem \Eq{eq:minp} can be regularized to enforce this symmetry.
    For a special case of a horizontal mirror symmetry, the regularizer could read
    \begin{equation*}
      \RegM(\Img) = \| \Img - \MirrorOp \Img\|_2,
    \end{equation*}
    where $\MirrorOp$ is a horizontal image ``flipping'' operator.
}

\end{document}